\def\year{2016}\relax
\documentclass[letterpaper]{article}
\usepackage{aaai17}
\usepackage{times}
\usepackage{latexsym}
\usepackage{times}
\usepackage{helvet}
\usepackage{courier}
\usepackage{amsfonts,amssymb}
\usepackage{amsmath}
\usepackage{graphicx}
\usepackage{algorithm}
\usepackage{algorithmic}
\usepackage{float}
\usepackage{array}
\usepackage{multirow}
\usepackage{url}

\title{SSP: Semantic Space Projection \\ for Knowledge Graph Embedding with Text Descriptions}
\date{}

\author{Han Xiao$^*$, Minlie Huang$^*$, Lian Meng, Xiaoyan Zhu \\
	State Key Lab. of Intelligent Technology and Systems, \\ 
	National Lab. for Information Science and Technology, \\
	Dept. of Computer Science and Technology, Tsinghua  University, Beijing 100084, PR China
	\\ bookman@vip.163.com; \{aihuang,zxy-dcs\}@tsinghua.edu.cn; mengl15@foxmail.com;
	\\$^*$Correspondence Authors: \url{http://www.ibookman.net}, \url{http://www.aihuang.org}
}

\begin{document}
\maketitle

\begin{abstract}
Knowledge graph embedding represents entities and relations in knowledge graph as low-dimensional, continuous vectors, and thus enables knowledge graph compatible with machine learning models. Though there have been a variety of models for knowledge graph embedding, most methods merely concentrate on the fact triples, while supplementary textual descriptions of entities and relations have not been fully employed. To this end, this paper proposes the \textbf{\textit{semantic space projection (SSP)}} model which jointly learns from the symbolic triples and textual descriptions. Our model builds interaction between the two information sources, and employs textual descriptions to discover semantic relevance and offer precise semantic embedding. Extensive experiments show that our method achieves  substantial improvements against baselines on the tasks of knowledge graph completion and entity classification. \textit{\textbf{Papers, Posters, Slides, Datasets and Codes: \url{http://www.ibookman.net/conference.html}}}
\end{abstract}

\section{Introduction}
Knowledge graph provides an effective basis for NLP in many tasks such as question answering, web search and semantic analysis. In order to provide a numerical computation framework for knowledge graph, knowledge graph embedding projects the entities and relations to a continuous low-dimensional vector space. More specifically, a fact in knowledge graph is usually represented as a symbolic triple ${(h, r, t)}$, while knowledge graph embedding attempts to represent the symbols with vectors, say \textbf{h,~r,~t}. To this end, a number of embedding methods have been proposed, such as TransE \cite{bordes2013translating}, PTransE \cite{lin2015modeling}, KG2E \cite{he2015learning}, etc.

\begin{figure}[H]
\centering
\includegraphics[width=1.0\linewidth]{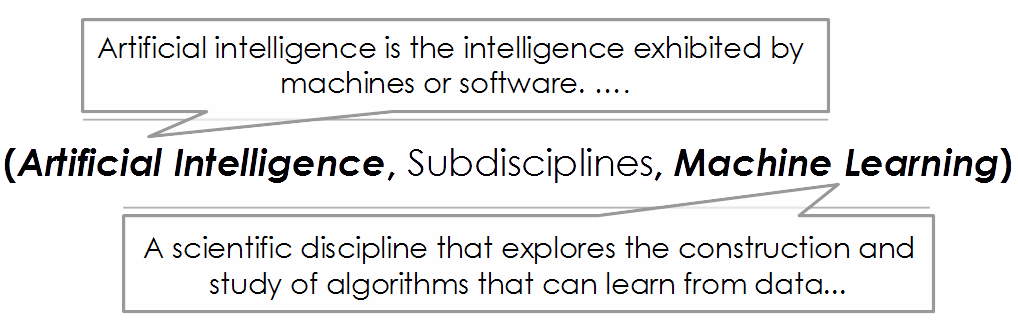}
\caption{Textual descriptions for entities in a fact triple.}
\label{fig:fig_1}
\end{figure}

As a key branch of embedding models, the translation-based methods adopt the principle of translating the head entity to the tail one by a relation-specific vector, or formally as $\mathbf{h+r=t}$. As Fig.\ref{fig:fig_1} shows, in the knowledge graph, the entities such as \textit{h, t} have textual descriptions, which contain much supplementary semantic information about knowledge triples.

Despite the success of conventional knowledge graph embedding models, there are still two reasons why textual descriptions would be necessary in this task: \textit{\textbf{discovering semantic relevance}} and \textit{\textbf{offering precise semantic expression}}. 

Firstly, the semantic relevance between entities is capable to recognize the true triples, which are difficult to be inferred only with fact triples. For example, the triple \textit{(Anna Roosevelt, Parents, Franklin Roosevelt)}, indicates \textit{``Franklin Roosevelt''} is the \textit{parent} of \textit{``Anna Roosevelt''}. However, it's quite difficult to infer this fact merely from other symbolic triples. In contrast, in the textual description of the head entity, there are many keywords such as \textit{``Roosevelt''} and \textit{``Daughter of the President''}, which may infer the fact triple easily. Specifically, we measure the possibility of a triple by projecting the loss onto a hyperplane that represents the semantic relevance between entities. Thus, it is always possible to accept a fact triple so long as the $l_2$-norm of the projected loss vector onto the semantic hyperplane is sufficiently small.

Secondly, precise semantic expression could promote the discriminative ability between two triples. For instance, when we query about the profession of \textit{``Daniel Sturgeon''}, there are two possible candidates: \textit{``politician''} and \textit{``lawyer''}. It's hard to distinguish if only focusing on the symbolic triples. However, the textual description of \textit{``Daniel Sturgeon''} is full of politics-related keywords such as \textit{``Democratic Party''}, \textit{``State Legislature''} etc. and even \textit{``Politician''}. The textual descriptions help to refine the topic of \textit{``Daniel Sturgeon''} in a more precise way from the social celebrities to the government officers, which makes the true answer \textit{``politician''} more preferable. Formally, even though the loss vectors of the two facts are almost of equal norm, 
after respectively projected onto the \textit{``politician''} and \textit{``lawyer''} related semantic hyperplanes, the losses are distinguished reasonably. In this way, precise semantic expression refines the embedding.

\begin{figure}[H]
\centering
\includegraphics[width=0.8\linewidth]{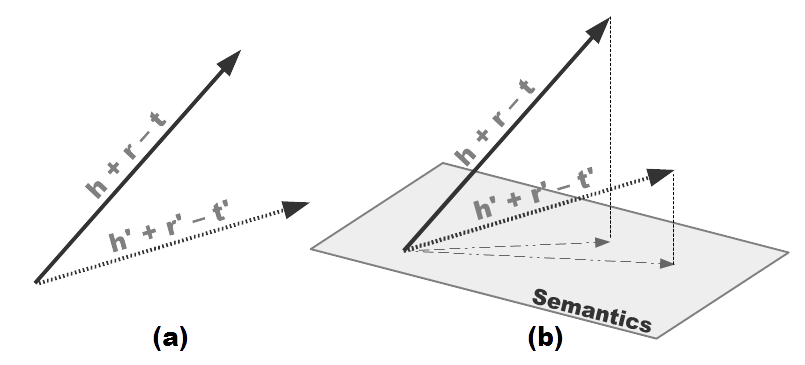}
\caption{Simple illustration of TransE and SSP where $\mathbf{h+r-t}$ is the loss vector. The loss vectors of the two triples in (a) are length-equal, thus, it is hard to identify the correctness.
In (b), 
we introduce a semantic hyperplane, and project the loss vectors to the hyperplane to consider the semantics of triples. }
\label{fig:fig_2}
\end{figure}

The existing embedding methods with textual semantics such as DKRL \cite{DKRL} and ``Jointly'' \cite{zhong2015aligning},
have achieved much success. But there is still an issue to be addressed, the \textit{\textbf{weak-correlation modeling}} issue that current models could hardly characterize the strong correlations between texts and triples. In DKRL, for a triple, the embedding vector of the head entity is translated to that of the tail one as well as possible, and then the vectors of texts and entities are concatenated as the final embedding vectors. While in the ``Jointly'' model the authors attempt to generate coherent embeddings of the corresponding entity and text. 
Both DKRL and ``Jointly'' apply first-order constraints which are weak in capturing the correlation of texts and triples.
\textbf{\textit{It's noteworthy that triple embedding is always the main procedure and textual descriptions must interact with triples for better embedding}}. Only in this way, the semantic effects could make more senses. 
For the above example of \textit{``Daniel Sturgeon''}, the textual descriptions imply two candidate answers \textit{``Banker''} and \textit{``Politician''}. Thus, by considering both triples and texts, we can obtain the true answer. Therefore, we focus on the stronger semantic interaction by projecting triple embedding onto a semantic subspace such as hyperplane, as shown in Fig.\ref{fig:fig_2}. Mathematically, the quadratic constraint is adopted to model the strong correlation, thus the embedding topologies are sufficiently semantics-specific.


We evaluate the effectiveness of our model \textit{Semantic Subspace Projection (SSP)} with two tasks on three benchmark datasets that are the subsets of Wordnet \cite{miller1995wordnet} and Freebase \cite{bollacker2008freebase}. Experimental results on the datasets show that our model consistently outperforms the other baselines with remarkable improvement.

\textbf{Contributions.} We propose a knowledge graph embedding method SSP which models the strong correlations between the symbolic triples and the textual descriptions by performing the embedding process in a semantic subspace. 
Besides, our method outperforms all the baselines on the tasks of knowledge graph completion and entity classification, which justifies the effectiveness of our proposal.


\section{Related Work}
We have surveyed the previous studies and categorized the embedding methods into two branches: \textit{Triple-only Embedding} models that use only symbolic triples and \textit{``Text-Aware'' Embedding} models that employ textual descriptions.

\subsection{Triple-only Embedding Models}
TransE \cite{bordes2013translating} is a pioneering work for this branch, which translates the head entity to the tail one by the relation vector, or $\mathbf{h+r=t}$. Naturally, the $L_2$ norm of the loss vector is the score function, which measures the plausibility of a triple and a smaller score is better. 

The following variants transform entities into different subspaces. ManifoldE \cite{ManifoldE} opens a classic branch, where a manifold is applied for the translation. TransH \cite{wang2014knowledge} utilizes the relation-specific hyperplane to lay the entities. TransR \cite{lin2015learning} applies the relation-related matrix to rotate the embedding space. Similar researches include TransG \cite{TransG}, TransD \cite{JiKnowledge}, \cite{GAKE} and TransM \cite{fan2014transition}.

Further researches encode additional structural information into embedding. PTransE \cite{lin2015modeling} is a path-based model, simultaneously considering the information and confidence level of a path in the knowledge graph. \cite{wang2015knowledge} incorporate the rules to restrict the embeddings for the complex relation types such as 1-N, N-1 and N-N. 
SSE \cite{guo2015semantically} aims at discovering the geometric structure of embedding topologies and then based on these assumptions, designs a semantically smoothing score function. Also, KG2E \cite{he2015learning} involves probabilistic analysis to characterize the uncertainty concepts of knowledge graph. There are also some other works such as SE \cite{bordes2011learning}, LFM \cite{jenatton2012latent}, NTN \cite{socher2013reasoning} and RESCAL \cite{nickel2011three}, TransA \cite{TransA}, etc.

\subsection{Text-aware Embedding Models} 
``Text-Aware'' Embedding, which attempts to representing knowledge graph with textual information, generally dates back to NTN \cite{socher2013reasoning}. NTN makes use of entity name and embeds an entity as the average word embedding vectors of the name. \cite{wang2014joint} attempts to aligning the knowledge graph with the corpus then jointly conducting knowledge embedding and word embedding. However, the necessity of the alignment information limits this method both in performance and practical applicability. Thus, \cite{zhong2015aligning} proposes the ``Jointly'' method that only aligns the freebase entity to the corresponding wiki-page. DKRL \cite{DKRL} extends the translation-based embedding methods from the triple-specific one to the ``Text-Aware'' model. More importantly, DKRL adopts a CNN-structure to represent words, which promotes the expressive ability of word semantics. Generally speaking, by jointly modeling knowledge and texts, \textit{text-aware embedding models} obtains the state-of-the-art performance.

\section{Methodology}
In this section, we first introduce our model and then provide two different perspective to address the advantages of our model. First of all, let us introduce some notations: all the symbols $h, t$ indicate the head and tail entity, respectively. $\mathbf{h}$ (or $\mathbf{t}$) is the embedding of the entity from the triples, $\mathbf{s_h}$ (or $\mathbf{s_t}$) is the semantic vector generated from the texts, and $d$ is the dimension of embedding.

The data involved in our model are the knowledge triples and the textual descriptions of entities. In experiments, we adopt the ``entity descriptions'' of Freebase and the textual definitions of Wordnet as textual information. 

\subsection{Model Description}
Previous analysis in the introduction suggests to characterize the strong correlation between triples and texts. For the purpose of interacting between the symbolic triples and textual descriptions, this paper attempts to restrict the embedding procedure of a specific triple in the semantic subspace. Specifically, we leverage a hyperplane with normal vector \fbox{$\mathbf{s} \doteq \mathcal{S}(\mathbf{s_h, s_t})$} as the subspace, where $\mathcal{S}:\mathbb{R}^{2d} \mapsto \mathbb{R}^d$ is the semantic composition function which will be discussed in the next section,
and $\mathbf{s_h, s_t}$ is the head-specific and tail-specific semantic vectors, respectively. 

The score function in the translation-based methods is $||\mathbf{h+r-t}||_2^2$, which means the triple embedding focuses on the loss vector \fbox{$\mathbf{e \doteq h+r-t}$}. According to our motivation, assuming $\mathbf{e}$ is length-fixed, the target is to maximize the component inside the hyperplane, which is $||\mathbf{e - s^\top e s}||_2^2 $. In detail, the component of the loss in the normal vector direction is $\mathbf{(s^\top e s)}$, then the other orthogonal one, that is inside the hyperplane, is $(\mathbf{e - s^\top e s})$.

It's natural that the norm of the loss vector should also be constrained. To this end, we introduce a factor $\lambda$ to balance the two parts, formally as:
$$ f_r(h,t) = - \lambda ||\mathbf{e - s^\top e s}||_2^2 + ||\mathbf{e}||_2^2$$
where $\lambda$ is a suitable hyper-parameter. Moreover, a smaller score means the triple is more plausible. For clarity, the definitions of the symbols are boxed. \textit{Notably, the projection part in our score function is negative, so smaller value means less loss.}

\subsection{Semantic Vector Generation}
 There are at least two methods that could be leveraged to generate the semantic vectors: topic model \cite{blei2012probabilistic} such as LSA, LDA, NMF \cite{stevens2012exploring} and word embedding such as CBOW \cite{mikolov2013linguistic}, Skip-Gram \cite{mikolov2013distributed}. More concretely, this paper adopts the topic model, treating each entity description as a document and then obtains the topic distribution of document as the semantic vector of entity. The entities are usually organized by the topic in knowledge base, for example, ``entity type'' is used in Freebase to categorize entities. Therefore, we conjecture that the topic model could be more suitable. Notably, the word embedding would also work well though maybe not better.

Given the pre-trained semantic vectors, our model fixes them and then optimizes the other parameters. We call this setting \textbf{Standard} (short as \textbf{Std.}). The reason why we could not adapt all the parameters, is that the training procedure would refill the semantic vectors and flush the semantics out. For the purpose of jointly learning the semantics and the embeddings, we conduct the topic model and the embedding model, simultaneously. In this way, the symbolic triples also impose a positive effect on the textual semantics and we call this setting \textbf{Joint}.
 
As each component of a semantic vector indicates the relevant level to a topic, we suggest the semantic composition should take the addition form: 
$$ \mathcal{S}\mathbf{(s_h, s_t)} = \frac{\mathbf{s_h + s_t}}{||\mathbf{s_h + s_t}||_2^2} $$
where the normalization is applied to make a normal vector. Since the largest components represent the topics, the addition operator corresponds to the union of topics, making the composition indicate the entire semantics. For example, when $\mathbf{s_h} = (0.1, 0.9, 0.0)$ and $\mathbf{s_t} = (0.8, 0.0, 0.2)$, the topic of the head entity is \#2 and that of the tail is \#1, while the composition is $\mathbf{s} = (0.45, 0.45, 0.10)$, corresponding to the topic of \#1,\#2, which is accordant to our intuition.

\subsection{Correlation Perspective}
Specifically, our model attempts to lay the loss $\mathbf{h' - t}$ onto the hyperplane, where $\mathbf{h' \doteq h + r}$ is the translated head entity. Mathematically, if a line lies on a hyperplane, so do all the points of this line. Correspondingly, the loss lays on the hyperplane, implying the head and tail entity also lay on it, as the beginning and ending point. \textbf{\textit{Thus, there exists the important restriction, that the entities co-occur in a triple should be embedded in the semantic space composed by the associated textual semantics}}. This restriction is implemented as a quadric form to characterize the strong correlation between texts and triples, in other words, to interact with both the information sources. A strong interaction between the textual descriptions and symbolic triples complements each other in a more semantics-specific form, which guarantees the semantic effects. More concretely, the embeddings are decided in the training procedure not only by triples but also by textual semantics, based on which, our embedding topologies are semantically different from the other methods. 

\subsection{Semantic Perspective}
There are two semantic effects for textual descriptions:  discovering semantic relevance and offering precise semantic expression. \textit{\textbf{Our model characterizes the strong correlations with a semantic hyperplane, which is capable of taking the advantages of both two semantic effects.}}

Firstly, according to the correlation perspective, the entities which are semantically relevant, approximately lay on a consistent hyperplane. Therefore, the loss vector between them ($\mathbf{h'-t}$) is also around the hyperplane. Based on this geometric insight, when a head entity matches a negative tail, the triple is far from the hyperplane, making a large loss to be classified. Conversely, even if a correct triple makes much loss, the score function after projected onto the hyperplane could be relatively smaller (or better). By this mean, the semantic relevance achieved from the texts, promotes embedding. For instance, the fact triple \textit{(Portsmouth Football Club, Locate, Portsmouth)} could hardly be inferred only within the triple embedding. It ranks 11,549 out of 14,951 by TransE in link prediction, which means a totally impossible fact. But the keywords ``Portsmouth'', ``England'', and ``Football'' occur many times in both the textual descriptions, making the two entities semantically relevant. Unsurprisingly, after the semantic projection, the case ranks 65 out of 14,951 in our model, which is much more plausible.

Secondly, all the equal-length loss vectors \footnote{Two vectors have equal-length iff they have the same $l_2$ norm.}
in TransE are equivalent in term of discrimination since the score function of TransE is $||\mathbf{h}+\mathbf{r}-\mathbf{t}||_2^2$, which leads to the weak distinction. However, with textual semantics, the discriminative ability could be strengthened in our model. Specifically, the equal-length loss vectors  are measured with the projection onto the corresponding semantic hyperplanes, which makes a reasonable division of the losses. For an instance of the query about which film \textit{``John Powell''} contributes to, there are two candidate entities, that the true answer \textit{``Kung Fu Panda''} and the negative one \textit{``Terminator Salvation''}. Without textual semantics, it's difficult to discriminate, thus the losses calculated by TransE are $8.1$ and $8.0$, respectively, leading to a hard decision. Diving into the textural semantics, we discover, \textit{``John Powell''} is much relevant to the topic of \textit{``Animated Films''}, which matches that of \textit{``Kung Fu Panda''} and does not for the other. Based on this fact, both the query and the true answer lie in the \textit{``Animated Films''}-directed hyperplane, whereas the query and the negative one do not co-occur in the corresponding associated semantic hyperplane. Thus, the projected loss of the true answer could be much less than that of the false one. Concretely, the losses in our model are $8.5$ and $10.8$, respectively, which are sufficient for discrimination. 


\subsection{Objectives \& Training} 
There are two parts in the objective function, which are respectively embedding-specific and topic-specific. To balance the two parts, a hyper-parameter $\mu$ is introduced. Overall, the total loss is:
\begin{eqnarray}
\mathcal{L} = \mathcal{L}_{embed} + \mu \mathcal{L}_{topic}
\end{eqnarray}
Notably, there is only the first part in the \textbf{Standard} setting where $\mu = 0$ in fact.

In term of the embedding-related objective, the rank-based hinge loss is applied, which means to maximize the discriminative margin between the golden triples and the negative ones:
\begin{eqnarray}
\mathop{\mathcal{L}} \limits_{embed} 
= \sum_{\tiny\mbox{$\begin{array}{c}
		(h,r,t) \in \Delta \\
		(h',r',t') \in \Delta'
		\end{array}$}
} [f_{r'}(h',t') - f_r(h,t) + \gamma]_{+} \nonumber
\end{eqnarray}
where $\Delta$ is the set of golden triples and $\Delta'$ is that of the negative ones. $\gamma$ is the margin, and $[~\cdot~]_+ = \max(~\cdot~, 0)$ is the hinge loss. The false triples are sampled with ``Bernoulli Sampling Method'' as introduced in \cite{wang2014knowledge} and the method selects the negative samples from the set of 
\begin{eqnarray}
\{(h', r, t) | h' \in E\} & \cup & \{(h, r, t') | t' \in E\} \nonumber \\ 
& \cup & \{(h, r', t) | r' \in R\} \nonumber
\end{eqnarray} 
We initialize the embedding vectors by the similar methods used in the deep neural network \cite{glorot2010understanding} and pre-train the topic model with Non-negative Matrix Factorization (NMF) \cite{stevens2012exploring}. The stochastic gradient descent algorithm (SGD) is adopted in the optimization.

For the topic-related objective, we take the advantage of the NMF Topic Model \cite{stevens2012exploring}, which is both simple and effective. Then we re-write the target as an inner-product form with the $L_2$-loss, stated as: 
\begin{eqnarray}
\mathcal{L}_{topic} = \sum_{e \in E,~w \in D_e} (C_{e,w} - \mathbf{s_e^\top w})^2 \\
\mathbf{s_e \ge 0}, \mathbf{w \ge 0}
\end{eqnarray}
where $E$ is the set of entities, and $D_e$ is the set of words in the description of entity $e$. $C_{e,w}$ is the times of the word $w$ occurring in the description $e$. $\mathbf{s_e}$ is the semantic vector of entity $e$ and $\mathbf{w}$ is the topic distribution of word $w$. Similarly, SGD is applied in the optimization.

Theoretically, our computation complexity is comparable to TransE, as $\mathcal{O}(\nu \times \mathcal{O}(TransE))$, and the small constant $\nu$ is caused by the projection operation and topic calculation. In practice, TransE costs 0.28s for one round in Link Prediction and our model costs 0.36s in the same setting. In general, TransE is most efficient among all the translation-based methods, while our method could be comparable to TransE in running time, justifying the efficiency of our model.

\section{Experiments}

\subsection{Datasets \& General Settings}  
Our experiments are conducted on three public benchmark datasets that are the subsets of Wordnet and Freebase. About the statistics of these datasets, we strongly suggest the readers to refer to \cite{DKRL} and \cite{lin2015learning}. The entity descriptions of FB15K and FB20K are the same as DKRL \cite{DKRL}, each of which is a small part of the corresponding wiki-page. The textual information of WN18 is the definitions that we extract from the Wordnet. Notably, for the zero-shot learning, FB20K is involved, which is also built by the authors of DKRL.



\subsection{Knowledge Graph Completion} 

\textbf{Evaluation Protocol.} The same protocol used in previous studies, is adopted. First, for each testing triple $(h,r,t)$, we replace the tail $t$ (or the head $h$) with every entity $e$ in the knowledge graph. Then, a probabilistic score of this corrupted triple is calculated with the score function $f_r(h,t)$. By ranking these scores in ascending order, we then get the rank of the original triple. The evaluation metrics are the average of the ranks as Mean Rank and the proportion of testing triple whose rank is not larger than 10 (as HITS@10). This is called ``Raw'' setting. When we filter out the corrupted triples that exist in the training, validation, or test datasets, this is the``Filter'' setting.  If a corrupted triple exists in the knowledge graph, ranking it ahead the original triple is also correct. To eliminate this effect, the ``Filter'' setting is more preferred. In both settings, a higher HITS@10 and a lower Mean Rank mean better performance.

\footnotetext[1]{This method involves much more extra text corpus, thus it's unfair to directly compare with others.}

\begin{table}[H]
	\centering
	\caption{Mean Rank and HITS@10 of Knowledge Graph Completion (For Predicting Entity) on FB15K and WN18.}
	\label{tab2}
	\renewcommand\arraystretch{1.05}
	\begin{tabular}{c|c|c|c|c}
		\hline \textbf{FB15K} & \multicolumn{2}{c|}{\textbf{Mean Rank}} & \multicolumn{2}{c}{\textbf{HITS@10}} \\ 
		\hline 
		\hline TransE & 210 & 119 & 48.5 & 66.1 \\
		\hline TransH  & 212 & 87 & 45.7 & 64.4 \\
		\hline Jointly & 167$~^1$ & 39$~^1$ & 51.7$~^1$ & 77.3$~^1$\\
		\hline DKRL(BOW) & 200 & 113 & 44.3 & 57.6 \\
		\hline DKRL(ALL) & \textbf{181} & 93 & 49.6 & 67.4 \\
		\hline
		\hline \textbf{SSP (Std.)} & 213 & 113 & 52.0 & 73.3 \\
		\hline \textbf{SSP (Joint)} & 188 & \textbf{85} & \textbf{53.5} & \textbf{77.1} \\
		\hline \hline
		\hline \textbf{WN18} & \multicolumn{2}{c|}{\textbf{Mean Rank}} & \multicolumn{2}{c}{\textbf{HITS@10}} \\ 
		\hline 
		\hline TransE & 263 & 251 & 75.4 & 89.2 \\
		\hline TransH & 401 & 338 & 73.0 & 82.3 \\
		\hline
		\hline \textbf{SSP (Std.)} & 312 & 193 & 81.3 & 91.4 \\
		\hline \textbf{SSP (Joint)} & \textbf{168} & \textbf{156} & \textbf{81.2} & \textbf{93.2} \\
		\hline
	\end{tabular} 
\end{table}

\textbf{Implementation.} As the datasets are the same, we directly reprint the experimental results of several baselines from the literature. We have attempted several settings on the validation dataset to get the best configuration. Under the ``bern.'' sampling strategy, the optimal configurations of our model SSP are as follows. For WN18, embedding dimension $d=100$, learning rate $\alpha=0.001$, margin $\gamma=6.0$, balance factor $\lambda=0.2$ and for SSP(Joint) $\mu=0.1$. For FB15K,  embedding dimension $d=100$, learning rate $\alpha=0.001$, margin $\gamma=1.8$, balance factor $\lambda=0.2$ and for SSP(Joint)  $\mu=0.1$. We train the model until convergence. Usually, it would converge until 10,000 rounds, but in current version, instead, we report the results of standard setting with 2,000 rounds. Regarding the joint setting, which is more difficult to converge, we train the model until 5,000 rounds.

\begin{table}[H]
	\centering
	\caption{Mean Rank and HITS@10 of Knowledge Graph Completion (For Predicting Relation) on FB15K.}
	\label{tab4}
	\renewcommand\arraystretch{1.05}
	\begin{tabular}{c|c|c|c|c}
		\hline \textbf{FB15K} & \multicolumn{2}{c|}{\textbf{Mean Rank}} & \multicolumn{2}{c}{\textbf{HITS@10}} \\ 
		\hline 
		\hline TransE & 2.91 & 2.53 & 69.5 & 90.2 \\
		\hline TransH & 8.25 & 7.91 & 60.3 & 72.5 \\
		\hline DKRL(BOW) & 2.85 & 2.51 & 65.3 & 82.7 \\
		\hline DKRL(ALL) & 2.41 & 2.03 & 69.8 & 90.8 \\
		\hline
		\hline \textbf{SSP (Std.)} & \textbf{1.58} & \textbf{1.22} & 69.9 & 89.2 \\
		\hline \textbf{SSP (Joint)} & 1.87 & 1.47 & \textbf{70.0} & \textbf{90.9} \\
		\hline
	\end{tabular} 
\end{table}

\textbf{Results} Evaluation results are reported in Tab.\ref{tab2} and Tab.\ref{tab4}. Note that ``Jointly'' refers to \cite{zhong2015aligning}. We observe that:
\begin{enumerate}
	\item SSP outperforms all the baselines in all the tasks, demonstrating the effectiveness of our models and the correctness of our theoretical analysis. Specifically, SSP(Joint) improves much more than SSP(Std.) for jointly learning the textual semantics and symbolic triples.
	\item DKRL and ``Jointly'' model only consider the first-order constraints, which interact between the textual and symbolic information, unsatisfactorily. By focusing on the strong correlation, SSP outperforms them. Notice that the ``Jointly'' model involves much more additional data to produce the result, but SSP also has an remarkable advantage against it. Though TransH is also a hyperplane-based method, SSP adopts the hyperplane in a semantics-specific way rather than a simple relation-specific form.  
	\item TransE could be treated as missing textual descriptions, and DKRL(BOW) could be viewed as missing symbolic triples. SSP (Joint) improves 12.4\% against TransE while 20.9\% against DKRL(BOW), illustrating the triple embedding is always the key point and the interactions between the  two information sources is a key factor.
\end{enumerate}

\subsection{Entity Classification} 
This task is essentially a multi-label classification, focusing on predicting entity types, which is crucial and widely used in many NLP \& IR tasks \cite{neelakantan2015inferring}. The entity in Freebase always has types, for instance, the entity types of \textit{``Scots''} are \textit{Human Language}, \textit{Rosetta Languoid}.

We adopt the same datasets as DKRL, for the details of which, we refer readers to \cite{DKRL}. Overall, this is a multi-label classification task with 50 classes, which means for each entity, the method should provide a set of types rather than a single type.

\textbf{Evaluation Protocol} In the training, we use the concatenation of semantic vector and embedding vector $\mathbf{(s_e,e)}$ as entity representation, which is the feature for the front-end classifier. For a fair comparison, our front-end classifier is also the Logistic Regression as DKRL in a one-versus-rest setting for multi-label classification. The evaluation is following \cite{neelakantan2015inferring}, which applies the mean average precision (MAP) that is commonly used in multi-label classification. Specifically, for an entity, if the methods predict a rank list of types and there are three correct types that lie in \#1, \#2, \#4, the MAP is calculated as $\frac{1/1+2/2+3/4}{3}$. For FB20K, the methods could only make use of the descriptions. Obviously, FB20K is a zero-shot scenario which is really hard.

\begin{table}
	\centering
	\caption{The MAP result of Entity Classification}
	\label{tab5}
	\renewcommand\arraystretch{1.15}
	\begin{tabular}{c|c|c}
		\hline \textbf{Metrics} & FB15K & FB20K \\
		\hline 
		\hline TransE & 87.8 & - \\
		\hline BOW & 86.3 & 57.5 \\
		\hline DKRL(BOW) & 89.3 & 52.0 \\
		\hline DKRL(ALL) & 90.1 & 61.9 \\
		\hline
		\hline NMF & 86.1 & 59.6 \\
		\hline \textbf{SSP (Std.)} & 93.2 & - \\
		\hline \textbf{SSP (Joint)} & \textbf{94.4} & \textbf{67.4} \\
		\hline
	\end{tabular} 
\end{table}

\textbf{Implementation.} As the datasets are the same, we directly report the experimental results of several baselines from the literature. We have attempted several settings on the validation dataset to get the best configuration. Under the ``bern.'' sampling strategy, the optimal configurations of our model SSP are as follows. For FB15K,embedding dimension $d=100$, learning rate $\alpha=0.001$, margin $\gamma=1.8$, balance factor $\lambda=0.2$ and for SSP(Joint) $\mu=0.1$. For FB20K, embedding dimension $d=100$, learning rate $\alpha=0.001$, margin $\gamma=1.8$, balance factor $\lambda=0.2$ and for SSP(Joint) $\mu=0.1$. We train the model until convergence.

\textbf{Results} The evaluation results are listed in Tab.~\ref{tab5}. Notably, TransE is only triple-specific and SSP (Std.) performs as well as the NMF in the zero-shot learning. 
Thus they are trivial for FB20K. 
The observations are as follows:
\begin{enumerate}
	\item Overall, our method SSP yields the best accuracy, justifying the effectiveness of our method.
	
	\item Compared to NMF and TransE, the results show that the interactions between triples and texts make a key difference and only one of them could hardly produce an effective performance. The promotion of SSP (Joint) in FB20K demonstrates the triple embedding also has a positive effect on textual semantics.
	\item Compared to TransE, the improvements illustrate the effectiveness of semantic integration. Compared to DKRL, the promotion demonstrates the important role of modeling strong correlation. 
\end{enumerate}

\begin{figure}
\centering
\includegraphics[width=0.7\linewidth]{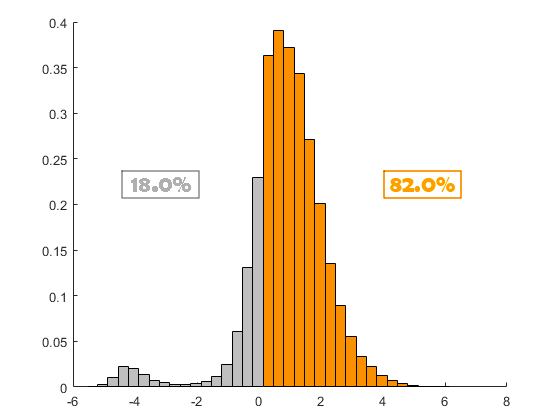}
\caption{Precise semantic expression analysis for SSP (Std.). The x-axis indicates the score difference, where a bigger value means better. The y-axis means the proportion of the corresponding triple pairs. The gray part indicates that both the TransE and SSP make mistakes while the yellow colored part means that SSP succeeds but TransE fails. The statistics are similar for the SSP(J.) setting.}
\label{fig:fig_3}
\end{figure}

\subsection{Semantic Relevance Analysis}
One benefit of modeling semantic relevance is the ability to correctly classify the triples that could not be discriminated by just using the information from symbolic triples. Hence, we make statistic analysis of the results in Link Prediction, as reported in the Tab.\ref{tab6}. 
The number in each cell means the number of triples whose rank is larger than $m$ in TransE and less than $n$ in our models. For instance, the number 601 means there are 601 triples whose ranks are less than 100 in SSP(S.) while their ranks are more than 500 in TransE.
%
Note that SSP(S.) indicates the standard setting, and SSP(J.) means the joint setting. The statistic results indicate that many triples benefit from the semantic relevance offered by textual descriptions. The experiments also justify the theoretical analysis about the semantic relevance and demonstrate the effectiveness of our models.

\begin{table}[H]
	\centering
	\caption{Rank statistics of Link Prediction. The number in each cell indicates the number of triples, and $r$ is the rank given by the corresponding models.\\ }
	\renewcommand\arraystretch{1.23}
	\label{tab6}
	\begin{tabular}{c|c|c}
 		 \hline ~ & $\mathbf{SSP(S.)_{\# r\le 100}}$ &  $\mathbf{SSP(J.)_{\# r\le 100}}$ \\
 		 \hline
 		 \hline $\mathbf{TransE_{\# r\ge 500}}$ & 601 & 672 \\
		 \hline $\mathbf{TransE_{\# r\ge 1000}}$ & 275 & 298 \\
		 \hline $\mathbf{TransE_{\# r\ge 2000}}$ & 80 & 89 \\
		 \hline $\mathbf{TransE_{\# r\ge 3000}}$ & 32 & 39 \\
		 \hline $\mathbf{TransE_{\# r\ge 5000}}$ & 3 & 3 \\
 		 \hline
	\end{tabular}
\end{table}

\subsection{Precise Semantic Expression Analysis}
As discussed previously, precise semantic expression leads to better discrimination. To justify this claim, we have collected the negative triples by link prediction, which are scored sightly better than the golden ones by TransE (that means, these are hard examples for TransE), and then plotted the SSP score difference between each corresponding pair of the negative and golden triples as Fig.\ref{fig:fig_3} shows. All these triples are predicted wrongly by TransE, but with precise semantic expression, our model correctly distinguishes $82.0\%$ (Std.) and $83.2\%$ (Joint) of them. In the histogram, the right bars indicate that SSP makes correct decision while TransE fails and the left bars mean both SSP and TransE fail. The experiments justify the theoretical analysis about the precise semantic expression and demonstrate the effectiveness of our models.


\section{Conclusion} 
In this paper, we propose the knowledge graph embedding model SSP, which jointly learns from the symbolic triples and textual descriptions. SSP could interact between the triples and texts by characterizing the strong correlations between fact triples and textual descriptions. The textual descriptions have much effect on discovering semantic relevance and then offering precise semantic expression. Extensive experiments show our method achieves the substantial improvements against the state-of-the-art baselines.

\textbf{Acknowledgement.} This work was partly supported by the National Basic Research Program (973 Program) under grant No. 2013CB329403, and the National Science Foundation of China under grant No.61272227/61332007.


\bibliographystyle{aaai}
\bibliography{SSP}

\begin{thebibliography}{}

\bibitem[\protect\citeauthoryear{Blei}{2012}]{blei2012probabilistic}
Blei, D.~M.
\newblock 2012.
\newblock Probabilistic topic models.
\newblock {\em Communications of the ACM} 55(4):77--84.

\bibitem[\protect\citeauthoryear{Bollacker \bgroup et al\mbox.\egroup
  }{2008}]{bollacker2008freebase}
Bollacker, K.; Evans, C.; Paritosh, P.; Sturge, T.; and Taylor, J.
\newblock 2008.
\newblock Freebase: a collaboratively created graph database for structuring
  human knowledge.
\newblock In {\em Proceedings of the 2008 ACM SIGMOD international conference
  on Management of data},  1247--1250.
\newblock ACM.

\bibitem[\protect\citeauthoryear{Bordes \bgroup et al\mbox.\egroup
  }{2011}]{bordes2011learning}
Bordes, A.; Weston, J.; Collobert, R.; Bengio, Y.; et~al.
\newblock 2011.
\newblock Learning structured embeddings of knowledge bases.
\newblock In {\em Proceedings of the Twenty-fifth AAAI Conference on Artificial
  Intelligence}.

\bibitem[\protect\citeauthoryear{Bordes \bgroup et al\mbox.\egroup
  }{2013}]{bordes2013translating}
Bordes, A.; Usunier, N.; Garcia-Duran, A.; Weston, J.; and Yakhnenko, O.
\newblock 2013.
\newblock Translating embeddings for modeling multi-relational data.
\newblock In {\em Advances in Neural Information Processing Systems},
  2787--2795.

\bibitem[\protect\citeauthoryear{Fan \bgroup et al\mbox.\egroup
  }{2014}]{fan2014transition}
Fan, M.; Zhou, Q.; Chang, E.; and Zheng, T.~F.
\newblock 2014.
\newblock Transition-based knowledge graph embedding with relational mapping
  properties.
\newblock In {\em Proceedings of the 28th Pacific Asia Conference on Language,
  Information, and Computation},  328--337.

\bibitem[\protect\citeauthoryear{Feng \bgroup et al\mbox.\egroup }{2016}]{GAKE}
Feng, J.; Huang, M.; Yang, Y.; and Zhu, X.
\newblock 2016.
\newblock {GAKE}: Graph aware knowledge embedding.
\newblock In {\em {COLING} 2016, 26th International Conference on Computational
  Linguistics}.

\bibitem[\protect\citeauthoryear{Glorot and
  Bengio}{2010}]{glorot2010understanding}
Glorot, X., and Bengio, Y.
\newblock 2010.
\newblock Understanding the difficulty of training deep feedforward neural
  networks.
\newblock In {\em International conference on artificial intelligence and
  statistics},  249--256.

\bibitem[\protect\citeauthoryear{Guo \bgroup et al\mbox.\egroup
  }{2015}]{guo2015semantically}
Guo, S.; Wang, Q.; Wang, B.; Wang, L.; and Guo, L.
\newblock 2015.
\newblock Semantically smooth knowledge graph embedding.
\newblock In {\em Proceedings of ACL}.

\bibitem[\protect\citeauthoryear{He \bgroup et al\mbox.\egroup
  }{2015}]{he2015learning}
He, S.; Liu, K.; Ji, G.; and Zhao, J.
\newblock 2015.
\newblock Learning to represent knowledge graphs with gaussian embedding.
\newblock In {\em Proceedings of the 24th ACM International on Conference on
  Information and Knowledge Management},  623--632.
\newblock ACM.

\bibitem[\protect\citeauthoryear{Jenatton \bgroup et al\mbox.\egroup
  }{2012}]{jenatton2012latent}
Jenatton, R.; Roux, N.~L.; Bordes, A.; and Obozinski, G.~R.
\newblock 2012.
\newblock A latent factor model for highly multi-relational data.
\newblock In {\em Advances in Neural Information Processing Systems},
  3167--3175.

\bibitem[\protect\citeauthoryear{Ji \bgroup et al\mbox.\egroup
  }{}]{JiKnowledge}
Ji, G.; He, S.; Xu, L.; Liu, K.; and Zhao, J.
\newblock Knowledge graph embedding via dynamic mapping matrix.

\bibitem[\protect\citeauthoryear{Lin \bgroup et al\mbox.\egroup
  }{2015}]{lin2015learning}
Lin, Y.; Liu, Z.; Sun, M.; Liu, Y.; and Zhu, X.
\newblock 2015.
\newblock Learning entity and relation embeddings for knowledge graph
  completion.
\newblock In {\em Proceedings of the Twenty-Ninth AAAI Conference on Artificial
  Intelligence}.

\bibitem[\protect\citeauthoryear{Lin, Liu, and Sun}{2015}]{lin2015modeling}
Lin, Y.; Liu, Z.; and Sun, M.
\newblock 2015.
\newblock Modeling relation paths for representation learning of knowledge
  bases.
\newblock {\em Proceedings of the 2015 Conference on Empirical Methods in
  Natural Language Processing (EMNLP). Association for Computational
  Linguistics}.

\bibitem[\protect\citeauthoryear{Mikolov \bgroup et al\mbox.\egroup
  }{2013}]{mikolov2013distributed}
Mikolov, T.; Sutskever, I.; Chen, K.; Corrado, G.~S.; and Dean, J.
\newblock 2013.
\newblock Distributed representations of words and phrases and their
  compositionality.
\newblock In {\em Advances in neural information processing systems},
  3111--3119.

\bibitem[\protect\citeauthoryear{Mikolov, Yih, and
  Zweig}{2013}]{mikolov2013linguistic}
Mikolov, T.; Yih, W.-t.; and Zweig, G.
\newblock 2013.
\newblock Linguistic regularities in continuous space word representations.
\newblock In {\em HLT-NAACL},  746--751.

\bibitem[\protect\citeauthoryear{Miller}{1995}]{miller1995wordnet}
Miller, G.~A.
\newblock 1995.
\newblock Wordnet: a lexical database for english.
\newblock {\em Communications of the ACM} 38(11):39--41.

\bibitem[\protect\citeauthoryear{Neelakantan and
  Chang}{2015}]{neelakantan2015inferring}
Neelakantan, A., and Chang, M.-W.
\newblock 2015.
\newblock Inferring missing entity type instances for knowledge base
  completion: New dataset and methods.
\newblock {\em arXiv preprint arXiv:1504.06658}.

\bibitem[\protect\citeauthoryear{Nickel, Tresp, and
  Kriegel}{2011}]{nickel2011three}
Nickel, M.; Tresp, V.; and Kriegel, H.-P.
\newblock 2011.
\newblock A three-way model for collective learning on multi-relational data.
\newblock In {\em Proceedings of the 28th international conference on machine
  learning (ICML-11)},  809--816.

\bibitem[\protect\citeauthoryear{Socher \bgroup et al\mbox.\egroup
  }{2013}]{socher2013reasoning}
Socher, R.; Chen, D.; Manning, C.~D.; and Ng, A.
\newblock 2013.
\newblock Reasoning with neural tensor networks for knowledge base completion.
\newblock In {\em Advances in Neural Information Processing Systems},
  926--934.

\bibitem[\protect\citeauthoryear{Stevens \bgroup et al\mbox.\egroup
  }{2012}]{stevens2012exploring}
Stevens, K.; Kegelmeyer, P.; Andrzejewski, D.; and Buttler, D.
\newblock 2012.
\newblock Exploring topic coherence over many models and many topics.
\newblock In {\em Proceedings of the 2012 Joint Conference on Empirical Methods
  in Natural Language Processing and Computational Natural Language Learning},
  952--961.
\newblock Association for Computational Linguistics.

\bibitem[\protect\citeauthoryear{Wang \bgroup et al\mbox.\egroup
  }{2014a}]{wang2014joint}
Wang, Z.; Zhang, J.; Feng, J.; and Chen, Z.
\newblock 2014a.
\newblock Knowledge graph and text jointly embedding.
\newblock In {\em EMNLP},  1591--1601.
\newblock Citeseer.

\bibitem[\protect\citeauthoryear{Wang \bgroup et al\mbox.\egroup
  }{2014b}]{wang2014knowledge}
Wang, Z.; Zhang, J.; Feng, J.; and Chen, Z.
\newblock 2014b.
\newblock Knowledge graph embedding by translating on hyperplanes.
\newblock In {\em Proceedings of the Twenty-Eighth AAAI Conference on
  Artificial Intelligence},  1112--1119.

\bibitem[\protect\citeauthoryear{Wang, Wang, and Guo}{2015}]{wang2015knowledge}
Wang, Q.; Wang, B.; and Guo, L.
\newblock 2015.
\newblock Knowledge base completion using embeddings and rules.
\newblock In {\em Proceedings of the 24th International Joint Conference on
  Artificial Intelligence}.

\bibitem[\protect\citeauthoryear{Xiao \bgroup et al\mbox.\egroup
  }{2015}]{TransA}
Xiao, H.; Huang, M.; Hao, Y.; and Zhu, X.
\newblock 2015.
\newblock {TransA}: An adaptive approach for knowledge graph embedding.
\newblock {\em arXiv preprint arXiv:1509.05490}.

\bibitem[\protect\citeauthoryear{Xiao, Huang, and Zhu}{2016a}]{ManifoldE}
Xiao, H.; Huang, M.; and Zhu, X.
\newblock 2016a.
\newblock From one point to a manifold: Knowledge graph embedding for precise
  link prediction.
\newblock In {\em Proceedings of the 25th International Joint Conference on
  Artificial Intelligence}.

\bibitem[\protect\citeauthoryear{Xiao, Huang, and Zhu}{2016b}]{TransG}
Xiao, H.; Huang, M.; and Zhu, X.
\newblock 2016b.
\newblock {TransG} : A generative model for knowledge graph embedding.
\newblock In {\em Proceedings of the 29th international conference on
  computational linguistics}.
\newblock Association for Computational Linguistics.

\bibitem[\protect\citeauthoryear{Xie \bgroup et al\mbox.\egroup }{2016}]{DKRL}
Xie, R.; Liu, Z.; Jia, J.; Luan, H.; and Sun, M.
\newblock 2016.
\newblock Representation learning of knowledge graphs with entity descriptions.

\bibitem[\protect\citeauthoryear{Zhong \bgroup et al\mbox.\egroup
  }{2015}]{zhong2015aligning}
Zhong, H.; Zhang, J.; Wang, Z.; Wan, H.; and Chen, Z.
\newblock 2015.
\newblock Aligning knowledge and text embeddings by entity descriptions.
\newblock In {\em Proceedings of EMNLP},  267--272.

\end{thebibliography}
\end{document}